\documentclass[11pt,letterpaper]{article}
\usepackage{emnlp2017}
\usepackage{times}
\usepackage{latexsym}
\usepackage{url}
\usepackage{mathtools}
\usepackage{amsmath}
\usepackage{algorithm}
\usepackage{lipsum}
\usepackage{cuted}
\usepackage[noend]{algpseudocode}
\usepackage{multirow}
\usepackage{graphicx}
\usepackage{enumitem}

\emnlpfinalcopy

\DeclareMathOperator*{\argmax}{arg\,max}

\def\emnlppaperid{398}

\title{Neural Machine Translation \\ Leveraging Phrase-based Models in a Hybrid Search}

\author{Leonard Dahlmann \and Evgeny Matusov \and Pavel Petrushkov \and Shahram Khadivi\\
  eBay Inc. \\
  Kasernenstr. 25 \\
  52064 Aachen, Germany \\
  {\tt \{fdahlmann, ematusov, ppetrushkov, skhadivi\}@ebay.com}\\
  }

\date{}

\begin{document}

\maketitle

\begin{abstract}
In this paper, we introduce a hybrid search for attention-based neural machine translation (NMT). A target phrase learned with statistical MT models extends a hypothesis in the NMT beam search when the attention of the NMT model focuses on the source words translated by this phrase. Phrases added in this way are scored with the NMT model, but also with SMT features including phrase-level translation probabilities and a target language model. Experimental results on German$\rightarrow$English news domain and English$\rightarrow$Russian e-commerce domain  translation tasks show that using phrase-based models in NMT search improves MT quality by up to 2.3\% BLEU absolute as compared to a strong NMT baseline.
\end{abstract}

\section{Introduction}
\label{sec:introduction}
Neural machine translation has become state-of-the-art in recent years, reaching higher translation quality than statistical phrase-based machine translation (PBMT) on many tasks. Human analysis~\cite{bentivogli2016neural} showed that NMT makes significantly fewer reordering errors, and also is able to select correct word forms more often than PBMT in the case of morphologically rich target languages. Overall, the fluency of the MT output improves when NMT is used, and the number of lexical choice errors is also reduced. However, state-of-the-art NMT approaches based on an encoder-decoder architecture with an attention mechanism as introduced by~\cite{bahdanau2014neural}
exhibit weaknesses that sometimes lead to MT errors which a phrase-based MT system does not make. In particular, PBMT usually can better translate rare words (e.g. singletons), as well as memorize and use phrasal translations. NMT has problems translating rare words because of limitations on the vocabulary size, as well as the fact that word embeddings are used to represent both source and target words. A rare word's embedding can not be trained reliably. %As a consequence, a rare target word can wrongly get a high probability in the NMT search just because its distance in the word embedding space is erroneously close to another unrelated word.

Another handicap of NMT is a general difficulty of fixing errors made by a neural MT system. 
Since NMT does not explicitly use or save word-to-word or phrase-to-phrase mappings, and its search is a target word beam search with almost no constraints, it is difficult to fix errors by an NMT system. 
It is important to quickly fix certain errors in real-life applications of MT systems to avoid negative user feedback or other (e.g. legal) consequences. An error identified in the output of a PBMT system can be fixed by tracing which phrase pair was used that resulted in the error, and down-weighting or even removing the phrase pair. Also, in PBMT it is easy to add an ``override'' translation.
%(e.g. using the exclusive XML markup mechanism in the popular Moses SMT decoder~\cite{koehn2007moses}), since at each step in the source cardinality-synchronous search as described e.g. by~\cite{zens2008improvements} it is always clear which source word range is translated next. All of this can not be easily done in the standard encoder-decoder NMT approach. NMT does not explicitly use or save word-to-word or phrase-to-phrase mappings, and its search is a target word beam search with at most a soft attention to source word positions not bound by any constraints.

In this work, we combine the strengths of NMT and PBMT approaches by introducing a novel hybrid search algorithm. In this algorithm, the standard NMT beam search is extended with phrase translation hypotheses from a statistical phrase table. The decision on when to use what phrasal translations is taken based on the attention mechanism of the NMT model, which provides a soft coverage of the source sentence words. All partial phrasal translations are scored with the NMT decoder and can be continued with a word-based NMT translation candidate or another phrasal translation candidate.

The proposed search algorithm uses a log-linear model in which the NMT translation score is combined with standard phrase translation scores, including a target $n$-gram language model (LM) score. Thus, a LM trained on additional monolingual data can be used. The decisions on the word order in the produced target translation are taken based only on the states of the NMT decoder.

This paper is structured as follows. We review related work in Section~\ref{sec:related-work}.
The baseline NMT model we use is described in Section~\ref{sec:background}, where we also recap the log-linear model combination used in PBMT.
Section~\ref{sec:hybrid-approach} presents the details of the proposed hybrid search. Experimental results are presented in Section~\ref{sec:experiments}, followed by conclusions and outlook in Section~\ref{sec:conclusion}.

\subsection{Related Work}
\label{sec:related-work}
In the line of research closely related to our approach, neural models are used as additional features in vanilla phrase-based systems. Examples include the work of~\cite{devlin2014fast},~\cite{junczys2016amu}, etc.
Such approaches have certain limitations: first, the search space of the model is still restricted by what can be produced using a phrase table extracted from parallel data based on word alignments. Second, the organization of the search, in which only a limited target word history (e.g. 4 last target words) is available for each partial hypothesis, makes it difficult to integrate recurrent neural network LMs and translation models which take all previously generated target words into account. That is why, for instance, the attention-based NMT models were usually applied only in rescoring~\cite{jtp16:iwslt}.

In~\cite{stahlberg2017mbr}, a two-step translation process is used, where in the first step a SMT translation lattice is generated, and in the second step the NMT decoder combines NMT scores with the Bayes-risk of the translations according to the lattice.  In contrast, we explicitly use phrasal translations and language model scores in an integrated search.

In~\cite{arthur2016incorporating}, a statistical word lexicon is used to influence NMT hypotheses, also based on the attention mechanism. \cite{gulcehre2015using} combine target $n$-gram LM scores with NMT scores to find the best translation. \cite{he2016improved} also use a target LM, but add further SMT features such as  word penalty and word lexica to the NMT beam search. To the best of our knowledge, no previous work extends the beam search with phrasal translation hypotheses of PBMT, like we propose in this paper.

In~\cite{tang2016neural}, the NMT decoder is modified to switch between using externally defined phrases and standard NMT word hypotheses. However, only one target phrase per source phrase is considered, and the reported improvements are significant only when manually selected phrase pairs (mostly for rare named entities) are used.

Somewhat related to our work is the concept of coverage-based NMT~\cite{tu2016modeling}, where the model architecture is changed to explicitly account for source coverage. In our work, we use a standard NMT architecture, but track coverage with accumulated attention weights.
%and ``hard'' word alignment of the additional phrase translation candidates. 

%Finally, we enhanced our baseline attention-based NMT model using subword units for vocabulary size reduction proposed by~\cite{sennrich2015neural}, stochastic optimization using the Adam algorithm~\cite{kingma2014adam} and dropout~\cite{gal2016theoretically}.

\section{Background}\label{sec:background}

\subsection{Neural MT}
\label{sec:neural-mt}
Neural MT proposed by \cite{bahdanau2014neural} maximizes the conditional log-likelihood of the target sentence $E:e_1,\dots,e_I$ given the source sentence $F:f_1,\dots,f_J$:
\begin{align}
	H_D = -\frac{1}{N} \sum_{n=1}^N \log p_{\theta}(E_n|F_n)
    %\label{eq: maximum-likelihood}
    \nonumber
\end{align}
where $(E_n, F_n)$ refers to the $n$-th training sentence pair in a dataset $D$, and $N$ denotes the total number of sentence pairs in the training corpus. When using the encoder-decoder architecture by \cite{cho2014learning}, the conditional probability can be written as: 
\vspace{-5mm}
\begin{align}
\begin{split}
	p(e_1 \cdots e_I | f_1 \cdots f_J) &= \prod_{i=1}^{I} p(e_i|e_{i-1} \cdots e_1, c)  
\end{split}
%\label{eq: enc-dec-prob}
 \nonumber
 \vspace{-5mm}
\end{align}
with
$ 
%\begin{align}
p(e_i|e_{i-1} \cdots e_1, c) = g(s_{i}, e_{i-1}, c) \nonumber
%\end{align}
$, 
where $I$ is the length of the target sentence and $J$ is the length of source sentence, $c$ is a fixed-length vector to encode the source sentence, $s_{i}$ is a hidden state of RNN at time step $i$, and $g(\cdot)$ is a non-linear function to approximate the word probability. When the attention mechanism is used, the vector $c$ in each sentence is replaced by a time-variant representation $c_i$ that is a weighted summary over a sequence of annotations $(h_1,\dots,h_{J})$, and $h_j$ contains information about the whole input sentence, but with a strong focus on the parts surrounding the $j$-th word~\cite{bahdanau2014neural}. Then, the context vector can be defined as:
\begin{align}
c_i = \sum_{j}^{J} \alpha_{ij} h_j
\hspace{5mm} \text{where} \hspace{5mm}
\alpha_{ij} = \frac{exp(r_{ij})}{\sum_{j=1}^{J} exp(r_{ij})}.
% \label{eq:ct}
 \nonumber
\end{align}
Therefore, $\alpha_{ij}$ is normalized over all source positions $j$.
Also, $r_{ij} = a(s_{i-1}, h_j)$ is the attention model used to calculate the log-likelihood of aligning the $i$-th target word to the $j$-th source word.
%
%score between previous state $s_{i-1}$ and the encoder state $h_j$. The score can be interpreted as the likelihood that the $i$-th target word is generated from the $j$-th word in the source sentence. 

\subsection{Phrase-based MT}
\label{sec:phrase-based-mt}
The log-linear model, as introduced in \cite{och2002loglinear}, allows decomposing the translation probability $Pr(e_1^I|f_1^J)$ by using an arbitrary number of features $h_m(f_1^J,e_1^I)$. Each feature is multiplied by a corresponding scaling factor $\lambda_m$:
\[
	Pr(e_1^I|f_1^J) = \frac{\exp\left(\sum_{m=1}^M\lambda_m h_m(f_1^J,e_1^I)\right)}{\sum_{\tilde{e}_1^{\tilde{I}}} \exp\left(\sum_{m=1}^M\lambda_m h_m(f_1^J,\tilde{e}_1^{\tilde{I}})\right)}.
\]

The standard PBMT approach uses a log-linear model in which bidirectional phrasal and lexical scores, language model scores, distortion scores, word penalties and phrase penalties are combined as features.

\section{Hybrid Approach}
\label{sec:hybrid-approach}
In this section we describe our proposed hybrid NMT approach. The algorithm allows translations to be generated partially by phrases\footnote{As in SMT, phrases can consist of only a single token.} and partially by words. Section \ref{sec:log-linear} describes the models we use to score hypotheses. The search algorithm is presented in Section \ref{sec:search}.

\subsection{Log-linear Combination}
\label{sec:log-linear}
We use a log-linear model combination to introduce SMT models into the NMT search. Since translations can be partially generated by phrases, we introduce the phrase segmentation $s_1^K$ as a hidden variable into the models similarly to \cite{zens2008improvements}, where $K$ is the number of phrases used in the translation. Note that, unlike standard PBMT, $s_1^K$ does not need to cover the whole source sentence, as parts of the translation can be generated by words. Using the maximum approximation, %the log-linear model then becomes
%\begin{equation}
%Pr(e_1^I|f_1^J) = \max_{s_1^K}\frac{\exp\left(\sum_{m=1}^M\lambda_m h_m(f_1^J,e_1^I,s_1^K)\right)}{\sum_{\tilde{e}_1^I,\tilde{s}_1^{\tilde{K}}} \exp\left(\sum_{m=1}^M\lambda_m h_m(f_1^J,\tilde{e}_1^I,\tilde{s}_1^{\tilde{K}})\right)},
%\end{equation}
the search criterion then is
\begin{equation}
\label{eq:loglin}
\hat{e}_1^{\hat{I}} = \argmax_{I,e_1^I} \left\{\max_{s_1^K} \sum_{m=1}^M \lambda_m h_m(f_1^J,e_1^I,s_1^K)\right\}.
\end{equation}
Let $\tilde{f}_k, \tilde{e}_k$ be the chosen phrase pairs in the segmentation $s_1^K$ for $k=1,\ldots,K$.
In our experiments with the proposed hybrid search, we use the following features:
\begin{enumerate}[noitemsep,topsep=0pt]
\item The NMT feature $h_\text{NMT}$.
	%\[
    %	h_\text{NMT}(f_1^J,e_1^I,s_1^K) = p_\theta(e_1^I|f_1^J).
    %\]
\item The word penalty feature $h_\text{WP}$ counts the number of target words.
	%\[
    %	h_\text{WP}(f_1^J,e_1^I,s_1^K) = I.
    %\]
    This feature can help control the length of translations.
\item The source word coverage feature $h_\text{SWC}$ counts the number of source words translated by phrases:
	\begingroup
    \setlength{\abovedisplayskip}{1pt}
	\setlength{\belowdisplayskip}{1pt}
	\[
    	h_\text{SWC}(f_1^J,e_1^I,s_1^K) = \sum_{k=1}^K |\tilde{f}_k|.
    \]
    \endgroup
    The purpose of this feature is to control the usage of phrases.
\item The phrase penalty feature $h_\text{PP}$ counts the number of phrases used.
	%\[
    %	h_\text{PP}(f_1^J,e_1^I,s_1^K) = K.
    %\]
    Together with the word penalty and the source word coverage feature, the phrase penalty can control the length of chosen phrases.
\item The $n$-gram language model feature $h_\text{LM}$.
	%\[
    %	h_\text{LM}(f_1^J,e_1^I,s_1^K) = \sum_{i=1}^I \log p(e_i|e_{i-n+1}^{i-1}).
    %\]
\item The bidirectional phrase features $h_\text{Phr}$ and $h_\text{iPhr}$.
	%\[
    %	h_\text{Phr}(f_1^J,e_1^I,s_1^K) = \sum_{k=1}^K \log p(\tilde{f}_k|\tilde{e}_k)
    %\]
    %\[
    %	h_\text{iPhr}(f_1^J,e_1^I,s_1^K) = \sum_{k=1}^K \log p(\tilde{e}_k|\tilde{f}_k)
    %\]
    Note that these features are only applied for those parts of the translation that are generated by phrases. The other parts get a phrase score of zero.
\end{enumerate}
The scaling factors $\lambda_m$ are tuned with minimum error rate training (MERT)~\cite{och2003mert} on $n$-best lists of the development set.

\subsection{Search}
\label{sec:search}

The algorithm is based on the beam search for NMT, which generates translations one word per time step in a left-to-right fashion.
We modify this search to allow hypothesizing phrases in addition to normal word hypotheses. The phrases are suggested based on the neural attention, starting from the source position with the maximal current attention. We only suggest phrases if a source position is focused. We check that suggested phrases do not overlap with already translated source words by keeping track of the sum of attention in previous time steps for each source position. Thus, the problem of global reordering is left entirely to the NMT model and we follow the attention when hypothesizing phrases.

Hypotheses are scored by NMT and SMT models. 
The beam is divided into two parts of fixed size: the word beam and the phrase beam.
The phrase beam is used to score target phrases which were hypothesized from an entry in a previous word beam.
In order to score a target phrase consisting of $k$ words with the NMT model, we use $k$ time steps, allowing us to keep the efficiency of batched NMT scoring.
Once a target phrase has been fully scored (and if the hypothesis has not been pruned), the hypothesis is returned to the word beam. Both beams are generated and pruned independently in each time step.

The algorithm has some hyper-parameters that need to be set manually. First, we have the beam size $N_p$ for phrase hypotheses and the beam size $N_w$ for word hypotheses. Second, $\tau_\text{focus}$ is the minimum attention that needs to be on a source position to consider it for extending with a phrase translation candidate whose source phrase starts on that position. Third, $\tau_\text{cov}$ is the minimum sum of attention of a source position over previous time steps at which it is considered to be covered. We do not hypothesize phrases that overlap with covered positions.

In the following, we describe the search in detail.
Let $f_1^J$ be the source sentence. Before search, we run the standard phrase matching algorithm on the source sentence to retrieve the translation options $E(j,j')$ for source positions $1 \le j < j' \le J$ from a given SMT phrase table.
With each hypothesis $h$, we associate the following items:
\begin{itemize}[noitemsep,topsep=0pt]
  \item $C(h,j)$ is the sum of the NMT attention to source position $j$ involved in generating the target words of $h$. This can be considered as a soft coverage vector for $h$.
  \item $Q(h)$ is the partial log-linear score of $h$ according to Equation \ref{eq:loglin}.
  \item $E(h)$ is the $n$-gram target word history of $h$.
  \item If $h$ is a phrase hypothesis with target phrase $\tilde{e}$, of which $k$ words already have been scored by NMT, then $P(h) \coloneqq (\tilde{e},k)$ is the phrase state.
\end{itemize}
Also, each hypothesis is associated with its corresponding NMT hidden state.
We initialize the beam to consist of an empty word hypothesis. Each step of the beam search proceeds as follows:

\begin{table*}[t]
\centering
\begin{tabular}{|c|c|c|c|c|c|}
\hline
\multicolumn{2}{|c|}{Data set}                     & \multicolumn{2}{c|}{WMT} & \multicolumn{2}{c|}{E-commerce} \\ \hline
\multicolumn{2}{|c|}{Language}                     & German       & English        & English          & Russian           \\ \hline
\multirow{3}{*}{Training} & Sentences              & \multicolumn{2}{|c|}{5,597,491}  & \multicolumn{2}{|c|}{2,919,406}           \\ \cline{2-6} 
                          & Running words          & 129,083,315       & 134,469,297       & 46,715,319          & 45,305,268          \\ \cline{2-6} 
                          & Full vocabulary        & 1,961,186    &    884,075     & 326,015           & 774,435           \\ \cline{1-6} 
\multirow{2}{*}{Dev}      & Sentences              & \multicolumn{2}{|c|}{2169 (WMT 15) }    & \multicolumn{2}{|c|}{950}             \\ \cline{2-6} 
                          & Running words          & 56,593        &  51,324      & 24,487            & 24,087            \\ \hline 
\multirow{2}{*}{Test}     & Sentences              & \multicolumn{2}{|c|}{6002 (WMT 14 + 16)}          & \multicolumn{2}{|c|}{1051 (item/product descriptions)}            \\ \cline{2-6} 
                          & Running words          & 160,469       &  144,387        & 29,165            & 26,476 \\ 
\hline
\end{tabular}
\caption{Corpus statistics for the WMT German$\rightarrow$English and e-commerce English$\rightarrow$Russian MT tasks.}
\label{table:corpus}
\end{table*}

\begin{enumerate}[noitemsep,topsep=0pt]
  \item Let $B = [B_w, B_p]$ be the previous beam with word/phrase hypotheses, respectively. First, we generate the attention vector $\alpha_{h,j}$ and the distribution over target words $\hat{p}_{h}(e)$ for each hypothesis $h \in B$ and word $e$ in the NMT target vocabulary $V_T$ using the NMT model in batched scoring \footnote{If a target word $e$ is not in $V_T$, set $\hat{p}_h(e)=\hat{p}_h(\text{UNK})$ where UNK is a special token denoting unknowns. Note that this almost never happens when using a word segmentation like BPE \cite{sennrich2015neural}.}.
  \item Initialize new beam $[B_{w}',B_{p}'] = [\emptyset, \emptyset]$.
  \item Generate new word hypotheses: find the maximal $N_w$ pairs $(h,e)$ with $h \in B_w$ and $e \in V_T$ according to the score $Q(h) + \lambda_{\text{NMT}} \cdot \log \hat{p}_{h}(e)$.
        For the top pairs $h' = (h,e)$, set
        \[
        \begin{split}
        &Q(h') = Q(h) + \lambda_{NMT} \cdot \log \hat{p}_{h}(e) \\
        &+ \lambda_{LM} \cdot \log p_{\text{LM}}(e|E(h)) + \lambda_{WP}
        \end{split}
        \]
        and insert $h'$ into $B_{w}'$.
        Update the soft coverage $C(h',j)=C(h,j) + \alpha_{h,j}$ for $1 \le j \le J$.
  \item Generate new phrase hypotheses: for each previous word hypothesis $h \in B_w$, convert the soft attention $C(h,\cdot)$ into a binary coverage set $C$, such that $j \in C$ iff. $C(h,j) > \tau_\text{cov}$.
        Identify the current NMT focus as \[\hat{j} = \argmax_{1 \le j \le J, \ \  \alpha_{h,j} > \tau_\text{focus}}{\alpha_{h,j}}. \]
        If there is no such $j$ with $\alpha_{h,j} > \tau_\text{focus}$, no phrase hypotheses are generated from $h$ in this step.
        Otherwise, for each source phrase length $l$ with $C \cap \{\hat{j}, \hat{j}+1, \ldots, \hat{j}+l-1\} = \emptyset$
        and each target phrase $\tilde{e} \in E(\hat{j},\hat{j}+l)$, create a new hypothesis $h'=(h,\tilde{e}_1)$ with the score
        \begin{equation}
        \begin{split}
          &Q(h') = Q(h) + \lambda_\text{NMT} \cdot \log \hat{p}_{h}(e_1) \\
          &+ \lambda_\text{LM} \cdot \log p_\text{LM}(\tilde{e}|E(h)) + |\tilde{e}| \cdot \lambda_\text{WP} \\
          &+ \lambda_\text{PP} + l \cdot \lambda_\text{SWC}.
        \end{split}
        \label{eq:phrasehyp}
        \end{equation}
        Note that, in this step, the full target phrase is scored using the language model, while only the first target word is scored using NMT.
        Initialize the phrase state of $h'$: $P(h')=(\tilde{e},1)$.
        As in step 3, update the soft coverage.
        If $|\tilde{e}|=1$, insert $h'$ into $B_w'$, otherwise insert into $B_p'$.
  \item Advance previous phrase hypotheses: for each $h \in B_p$, with phrase state $P(h)=(\tilde{e},k)$, score the $(k+1)$-th target word of $\tilde{e}$ using NMT, setting $h'=(h,\tilde{e}_{k+1})$ and \[ Q(h')=Q(h)+\lambda_\text{NMT}\cdot\log \hat{p}_{h}(\tilde{e}_{k+1}). \]
       As in step 3, update the soft coverage.
       Set the new phrase state as $P(h')=(\tilde{e},k+1)$.
        If $k+1=|\tilde{e}|$, we are finished scoring the phrase and $h'$ is inserted into $B_w'$. Otherwise, $h'$ is inserted in $B_p'$.
  \item Prune $B_w'$ to $N_w$ entries and $B_p'$ to $N_p$ entries according to $Q(\cdot)$. 
%  If there are fewer $newPhraseHyps$ than $beamSizePhrases$, fill the beam up with word hypotheses.
  \item Insert all hypotheses from the pruned $B_w'$ and $B_p'$ where the last word is the sentence end token into the set of finished hypotheses $B_f$.
  \item ${B := [B_w', B_p']}$.
\end{enumerate}
If phrase scores from a phrase table are to be included in the search, Equation \ref{eq:phrasehyp} needs to be modified by adding $\lambda_\text{Phr} \log p(\tilde{f}|\tilde{e})$ and $\lambda_\text{iPhr} \log p(\tilde{e}|\tilde{f})$.

As in the pure NMT beam search, this procedure is repeated until either the last word of all hypotheses in a step is the sentence end token, or $2\cdot J$ many beam steps have been performed. Finally, the best translation is chosen as the one in $B_f$ with the highest score.

Note that the same target sequence can be generated with different phrasal segmentations. During search, if two hypotheses have the same full target history in a beam, we recombine them and discard the hypothesis with the lower score.

\section{Experiments}\label{sec:experiments}
We perform experiments comparing the translation quality of our hybrid approach to phrase-based and pure end-to-end NMT baselines. We present results on two tasks: an in-house English$\rightarrow$Russian e-commerce task (translation of real product/item descriptions from an e-commerce site), and the WMT 2016 German$\rightarrow$English task (news domain). The corpus statistics are shown in Table \ref{table:corpus}. 

\begin{table*}
\centering
\begin{tabular}{|l|c|c|c|c|c|}
\hline
     & & \multicolumn{2}{c|}{Item descriptions} & \multicolumn{2}{c|}{Product descriptions} \\
\hline
System description & Beam size & BLEU [\%] & TER [\%] & BLEU [\%] & TER [\%] \\
\hline
Phrase-based & - & 21.3 & 61.6 & 22.7 & 56.6 \\
+ $1000$-best rescoring with NMT & - & 23.1 & 60.1 & 25.8 & 54.7 \\
\hline
NMT & 12 & 26.4 & 56.4 & 28.4 & 52.0 \\
NMT & 128 & 26.3 & 56.6 & 28.5 & 51.9 \\
\hline
Full hybrid approach & 128 & 26.7 & 56.1 & 29.9 & 51.2 \\
+ extra LM data & 128 & 27.4 & 55.4 & 30.8 & 50.5 \\
\hline
NMT + WP + LM (with extra data) & 128 & 26.2 & 57.3 & 29.0 & 51.8 \\
\hline
\end{tabular}
\caption{Overview of translation results on the e-commerce English$\rightarrow$Russian task.}
\label{table:enru}
\end{table*}

For the English$\rightarrow$Russian task, the parallel training data consists of an in-domain part (ca. 5.5M running words) of product/item titles and descriptions and other e-commerce content.  
The rest is out-of-domain data (UN, subtitles, TAUS data collections, etc.) sampled to have significant $n$-gram overlap with the in-domain description data.
Item descriptions are provided by private sellers and, like any user-generated content, may contain ungrammatical sentences, spelling errors, and other noise. Product descriptions usually originate from product catalogs and are more ``clean'', but on the other hand, are difficult to translate because of rare domain-specific terminology. 
Both types of text contain itemizations, measurement units, and other structures which are usually not found in normal sentences.
We tune the system on a development set that is a mix of product and item descriptions, and evaluate on separate product/item description test sets.  For development and test sets, two reference translations are used. 

The German$\rightarrow$English system is trained on parallel corpora provided for the constrained WMT 2017 evaluation (Europarl, Common Crawl, and others). We use the WMT 2015 evaluation data as development set, and the evaluation is performed on two sets from the WMT evaluations in 2014 and 2016. Only a single human reference translation is provided. 

For the phrase-based baselines, we use an in-house phrase-decoder~\cite{iwslt10:EC:apptek} which is similar to the Moses decoder \cite{koehn2007moses}. We use standard SMT features, including word-level and phrase-level translation probabilities, the distortion model, 5-gram LMs, and a 7-gram joint translation and reordering model reimplemented based on the work of~\cite{guta15jtr}. The language model for the e-commerce task is trained on additional monolingual Russian item description data containing 28.2M words. For the WMT task, we use the English News Crawl data containing 3.8B words for additional language model data. The tuning is performed using MERT~\cite{och2003mert} to increase the BLEU score on the development set. To stabilize the optimization on the English$\rightarrow$Russian task, we detach Russian morphological suffixes from the word stems both in hypotheses and references using a context-independent ``poor man's'' morphological analysis. We prefix each suffix with a special symbol and treat them as separate tokens. 

We have implemented our NMT model in Python using the TensorFlow\footnote{\url{http://tensorflow.org}} deep learning library. We use the embedding size of 620, RNN size of 1000 and GRU cells.
The model is trained with maximum likelihood loss for 15 epochs using Adam optimizer \cite{kingma2014adam} on complete data in batches of 100 sentences. The learning rate is initialized to 0.0002, decaying by 0.9 each epoch. For regularization we use L2 loss with weight $10^{-7}$ and dropout following~\citet{gal2016theoretically}. We set the dropout probability for input and recurrent connections of the RNN to 0.2 and word embedding dropout probability to 0.1.
On the English$\rightarrow$Russian task, the model is then fine-tuned on in-domain data for 10 epochs. The vocabulary is limited using byte pair encoding (BPE) \cite{sennrich2015neural} with 40K splits separately for each language. To speed up training we use approximate loss as described in~\cite{DBLP:conf/acl/JeanCMB15}.
For pure NMT experiments, we employ length normalization \cite{google2016nmt}, as otherwise short translations would be favored.

\begin{table*}
\centering
\begin{tabular}{|l|c|c|c|c|}
\hline
     & \multicolumn{2}{c|}{Item descriptions} & \multicolumn{2}{c|}{Product descriptions} \\
\hline
System description & BLEU [\%] & TER [\%] & BLEU [\%] & TER [\%] \\
\hline
Full hybrid approach & 27.4 & 55.4 & 30.8 & 50.5 \\
\hline
Without LM            & 26.5 & 55.9 & 29.2 & 51.0 \\
Without source word coverage feature & 26.7 & 56.1 & 29.4 & 51.2 \\
Without phrase scores & 27.2 & 55.9 & 30.6 & 50.6 \\
Maximal source phrase length 1 & 26.7 & 56.4 & 29.1 & 51.6 \\
Minimal source phrase length 2 & 27.0 & 55.9 & 30.0 & 51.1 \\
\hline
\end{tabular}
\caption{Translation results of the hybrid approach on the e-commerce English$\rightarrow$Russian task with different SMT model combinations. The first row shows results with all models enabled. In the following rows, we either remove or limit exactly one model compared to the full system.}
\label{table:enruhybrid}
\end{table*}

For the hybrid approach, we use the same trained end-to-end model as in the NMT baseline. We use all the phrase-based model features plus the NMT score and run MERT as described in Section \ref{sec:log-linear}. Language models are trained on the level of BPE tokens.
We consider at most 100 translation options for each source phrase.
If not specified otherwise, we use a beam size of 96 for phrase hypotheses and a beam size of 32 for word hypotheses, resulting in a combined beam size of 128. Furthermore, we set the focus threshold $\tau_\text{focus}=0.3$ and the coverage threshold $\tau_\text{cov}=0.7$ by default. We also perform experiments where these hyper-parameters are varied.

\subsection{E-commerce English$\rightarrow$Russian}
The results on the e-commerce English$\rightarrow$Russian task are summarized in Table \ref{table:enru}. 
\subsubsection*{NMT vs. phrase-based SMT}
The pure NMT system exhibits large improvements over the phrase-based baseline\footnote{The significance of these improvements
was also confirmed by an in-house human evaluation with 3 judges.}.
These improvements are also significantly larger than when we use the NMT model to rescore PBMT 1000-best lists. NMT results are not improved when the beam size is increased from 12 to 128. 

\subsubsection*{Hybrid search vs. pure NMT search}
For the hybrid approach, we train a phrase-table on the in-domain data and split the source and target phrases with BPE afterwards for compatibility with the NMT vocabulary.
With the hybrid approach, when using a LM trained only on the target side of bilingual data, we get an improvement of 0.3\% BLEU on item descriptions and 1.4\% BLEU on product descriptions over the pure NMT system. When we use the LM trained on extra monolingual data, we get total improvements of 1.0\% BLEU and 2.3\% BLEU with the hybrid approach. In contrast, when we add this language model and a word penalty on top of the pure NMT system and tune scaling factors with MERT, we get small improvements (last row of Table \ref{table:enru}) only on product descriptions. This shows that the hybrid approach can exploit the LM better than a purely word-based NMT approach.
We have also performed experiments utilizing the additional monolingual data for synthetic training data for NMT as in \cite{rico2015synthetic}, but did not get improvements.

To analyze the improvements of the hybrid system, we perform experiments in which we either disable or limit some of the SMT models. The results are shown in Table \ref{table:enruhybrid}. Without the language model, the hybrid approach has almost no improvements over the NMT baseline. This indicates that the language model is crucial in selecting appropriate phrase candidates. Similarly, when we disable the source word coverage feature, the translation quality is degraded, suggesting that this feature helps choose between phrase hypotheses and word hypotheses during the search. Next, we do not use phrase-level scores. Here, we observe only a small degradation of translation quality. Finally, we limit the source length of phrases used in the search, allowing only one-word source phrases in one experiment and only source phrases with two or more words in another experiment. In both cases, the translation quality decreases. Thus, both one-word phrases and longer phrases are necessary to obtain the best results.

\subsubsection*{Tuning the beam size}
Next, we study the effect of different beam sizes on translation quality.
The results are shown in Table \ref{table:enrubeamsize}. Note that we retune the system for each choice. With a total beam size of 128, we get the best results by using a phrase beam size of 96 and a word beam size of 32. When we use a phrase beam size of 116 or 64 instead, the translation quality worsens.
%This shows that choosing the ratio between phrase and word beam sizes can be critical.
In another experiment, we decrease the total beam size to 64. The translation quality degrades only slightly, which means that we can still expect MT quality improvements with hybrid search even if we optimize the system for speed. To further test this, we reduce the beam sizes to $N_w=12$ and $N_p=4$ after tuning with $N_w=32$ and $N_p=96$. We get BLEU scores of 27.1\% on item descriptions and 30.1\% on product descriptions, losing 0.3\% and 0.7\% BLEU respectively compared to the full beam size.

\begin{table}[th]
\centering
\begin{tabular}{|c|c|c|c|c|c|}
\hline
\multicolumn{2}{|c|}{Beam size} & \multicolumn{2}{c|}{Item descr.} & \multicolumn{2}{c|}{Product descr.} \\
\hline
$N_p$ & $N_w$ & BLEU & TER & BLEU & TER \\
 & & [\%] & [\%] & [\%] & [\%] \\
\hline
116 & 12 & 26.7 & 55.9 & 29.8 & 51.1 \\
96 & 32 & 27.4 & 55.4 & 30.8 & 50.5 \\
64 & 64 & 26.8 & 55.6 & 30.1 & 50.7 \\
32 & 32 & 27.1 & 55.8 & 30.7 & 50.5 \\
\hline
\end{tabular}
\caption{Effect of the beam size (word beam size $N_w$ + phrase beam size $N_p$) for the hybrid approach on the e-commerce English$\rightarrow$Russian task.}
\label{table:enrubeamsize}
\end{table}

\subsubsection*{Tuning the attention focus/coverage thresholds}

Table \ref{table:enruthresh} shows results with different values for the coverage threshold $\tau_\text{cov}$. Again, we retune the system for each choice. Setting the coverage threshold to 1.0 or even disabling the coverage check (by setting $\tau_\text{cov}=\infty$) has little effect on the translation scores on this task. This can be explained by the fact that translation from English to Russian is mostly monotonic. We also tried varying the focus threshold $\tau_\text{focus}$ between $0.0$ and $0.3$ but did not notice any significant effect on this task.
\begin{table}[th]
\centering
\begin{tabular}{|c|c|c|c|c|c|}
\hline
\multicolumn{2}{|c|}{} & \multicolumn{2}{c|}{Item descr.} & \multicolumn{2}{c|}{Product descr.} \\
\hline
$\tau_\text{focus}$ & $\tau_\text{cov}$ & BLEU & TER & BLEU  & TER \\
 &  & [\%] & [\%] & [\%] & [\%] \\
\hline
0.3 & 0.7 & 27.4 & 55.4 & 30.8 & 50.5 \\
%0.0 & 0.7 & 26.5 & 56.1 & 29.8 & 51.4 \\
0.3 & 1.0 & 27.2 & 55.4 & 30.3 & 50.3 \\
0.3 & $\infty$ & 27.5 & 55.4 & 30.4 & 50.9 \\
\hline
\end{tabular}
\caption{Effect of the threshold parameters on the hybrid approach on the e-commerce English$\rightarrow$Russian task.}
\label{table:enruthresh}
\end{table}

\begin{table*}[th]
\centering
\begin{tabular}{|l|c|c|c|c|c|}
\hline
     & & \multicolumn{2}{c|}{newstest2014} & \multicolumn{2}{c|}{newstest2016} \\
\hline
System description & Beam size & BLEU [\%] & TER [\%] & BLEU [\%] & TER [\%] \\
\hline
Phrase-based & - & 22.9 & 59.4 & 26.9 & 54.1 \\
+ News Crawl LM data & - & 25.4 & 59.0 & 29.2 & 53.8 \\
\hline
NMT & 12 & 26.9 & 53.0 & 32.3 & 47.6 \\
NMT & 64 & 27.0 & 53.0 & 32.2 & 47.6\\
\hline
Hybrid approach & 64 & 27.8 & 53.2 & 32.4 & 48.2 \\
+ tuning $\tau_\text{focus}$, $\tau_\text{cov}$ & 64 & 28.0 & 53.0 & 33.3 & 47.4 \\
+ News Crawl LM data & 64 & 29.7 & 52.2 & 35.3 & 46.7 \\
\hline
\end{tabular}
\caption{Overview of translation results on the WMT German$\rightarrow$English task.}
\label{table:wmt}
\end{table*}

\subsubsection*{Analysis}
To understand the behavior of the hybrid search, we count the number of source words that are translated by phrases in the product descriptions test set. Of the 9320 source words, 7109 (76.3\%) are covered by phrase hypotheses. 78.3\% of the source phrases are unigrams, 19.5\% are bigrams and 2.2\% are trigrams or longer.
Among the many one-word phrases used, almost all (99.2\%) are also within the top 3 predictions of word-based NMT, and 90.3\% are equal to the top NMT prediction.
%This means that the information that a source/target word pair is contained in a phrase table by itself already reinforces identical NMT hypotheses, guiding NMT to better translations. 

Further human analysis by a native Russian speaker of the pure NMT vs. hybrid search translations shows that hybrid search is often able to correct the following known NMT handicaps:
\begin{itemize}[noitemsep,topsep=0pt]
\item incorrect translation of rare words (among other reasons, due to incorrect sub-word unit translation in which rare words are aggressively segmented).

\item repetition of same or similar words as a result of multiple attention to the same source word, as well as untranslated words that received no attention.

\item incorrect or partially correct word-by-word translation when a phrasal (non-literal) translation should be used instead.

\end{itemize}
In all of these cases, the usage of phrasal translations is able to better enforce the coverage, and this, in turn, leads to improved lexical choice. The fact that not many long phrase pairs are selected indicates, in our opinion, that the search and modeling problem in NMT is far from being solved: with the right, diverse model scores, the proposed hybrid search is able to select and extend better hypotheses with words, most of which already had a high NMT probability. Yet they are not always selected in the pure NMT beam search, among other reasons, due to competition from words erroneously placed near them in the embedding space.
%Yet they are not always selected in the pure NMT beam search, among other reasons, due to the wrong attention and competition from words erroneously placed near them in the embedding space.

\subsection{WMT 2016 German$\rightarrow$English}
The results on the WMT German$\rightarrow$English task are shown in Table \ref{table:wmt}. The initial phrase-based baseline uses the 5-gram language model estimated on the target side of bilingual data. By adding the News Crawl LM data, we gain 2.5\% and 2.3\% BLEU on the test sets, but PBMT still is behind NMT.

For the hybrid approach, we use a beam size of $64$ and a maximal number of beam steps of $1.5 \cdot J$ (instead of $2 \cdot J$) to speed up experiments. We use separate word penalty features, one for word-based hypotheses and one for phrase-based hypotheses to allow for more control of translation lengths.
With the hybrid approach, using the 5-gram language model estimated on the target side of bilingual data, and phrase scores, we get small improvements in BLEU over the NMT baseline. However, the TER increases. We experiment with different thresholds, setting $\tau_\text{focus}=0.1$ and $\tau_\text{cov}=1.0$. With this hybrid system, we get improvements of 1.0\% and 1.1\% BLEU over pure NMT. Finally, we add the News Crawl LM data on top. This significantly improves the results by 1.7\% and 2.0\% BLEU. In total, we gain 2.7\% and 3.1\% BLEU over pure NMT. These results reinforce the fact that, similar to PBMT, language model quality is important for the proposed hybrid search. In contrast, we have also tried applying only the LM (including News Crawl data) with a word penalty on top of NMT, but did not get consistent improvements.

\begin{figure}[ht]
\centering
\includegraphics[width=.42\textwidth]{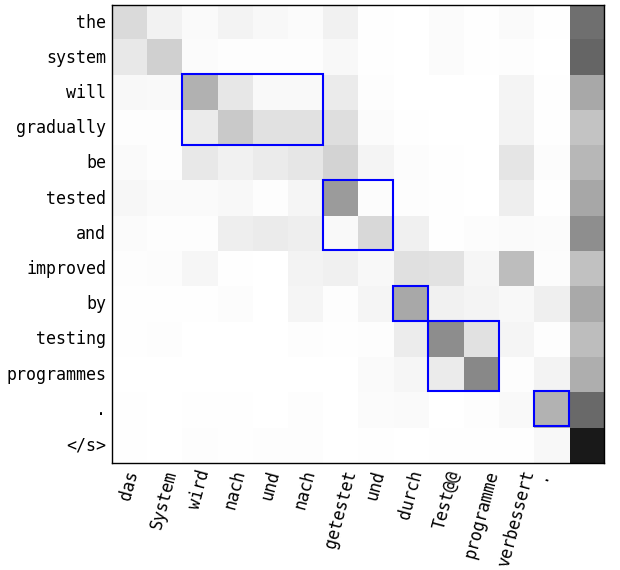}
\caption{Example alignment from the hybrid search, with the source sentence on the bottom and the translation on the left. The blue rectangles signify phrase pairs on top of the NMT attention. The pure NMT translation is ``the system is tested after and after testing and improved by testing programs.''}
\end{figure}

Figure 1 shows an example for the phrase pairs chosen by the hybrid system on top of the NMT attention. The hybrid approach correctly translates the German idiom ``nach und nach'' as ``gradually'', while the pure NMT system incorrectly translates it word-by-word as ``after and after''.

\section{Conclusion}\label{sec:conclusion}
In this work, we proposed a novel hybrid search that extends NMT with phrase-based models. The NMT beam search was modified to insert phrasal translations based on the current and accumulated attention weights of the NMT decoder RNN. The NMT model score was used in a log-linear model with standard phrase-based scores as well as an $n$-gram language model. We described the algorithm in detail, in which we keep separate beams for NMT word hypotheses and hypotheses with an incomplete phrasal translation, as well as introduce parameters which control the source sentence coverage. Numerous experiments on two large vocabulary translation tasks showed that the hybrid search improves BLEU scores significantly as compared to a strong NMT baseline that already outperforms phrase-based SMT by a large margin.

In the future, we plan to focus on integration of phrasal components into NMT training, including better coverage constraints, as well as methods for context-dependent translation override within our hybrid search algorithm.

\section*{Acknowledgments}
We would like to thank Tamer Alkhouli and Jan-Thorsten Peter for helpful discussions. We thank the anonymous reviewers for their suggestions.

% Do not number the acknowledgment section.

\bibliography{emnlp2017}
\bibliographystyle{emnlp_natbib}

\end{document}